\documentclass[10pt,twocolumn,letterpaper]{article}

\usepackage{cvpr}
\usepackage{times}
\usepackage{epsfig}
\usepackage{graphicx}
\usepackage{amsmath}
\usepackage{amssymb}
\usepackage{multirow}
\usepackage{booktabs}
\usepackage{array}
\usepackage{authblk}
\usepackage{footnote}
\usepackage{bm}
\makesavenoteenv{tabular}

\usepackage[breaklinks=true,bookmarks=false]{hyperref}

\cvprfinalcopy 

\def\cvprPaperID{****} 

\begin{document}

\title{The Solution for Language-Enhanced Image New Category Discovery}
\author[1]{Haonan Xu}
\author[1]{Dian Chao}
\author[1]{Xiangyu Wu}
\author[1]{Zhonghua Wan}
\author[1]{{Yang Yang\thanks{Corresponding author: Zhonghua Wan}}}
\affil[1]{Nanjing University of Science and Technology}

\maketitle
\begin{abstract}
Treating texts as images, combining prompts with textual labels for prompt tuning, and leveraging the alignment properties of CLIP have been successfully applied in zero-shot multi-label image recognition. Nonetheless, relying solely on textual labels to store visual information is insufficient for representing the diversity of visual objects. In this paper, we propose reversing the training process of CLIP and introducing the concept of Pseudo Visual Prompts. These prompts are initialized for each object category and pre-trained on large-scale, low-cost sentence data generated by large language models. This process mines the aligned visual information in CLIP and stores it in class-specific visual prompts. We then employ contrastive learning to transfer the stored visual information to the textual labels, enhancing their visual representation capacity. Additionally, we introduce a dual-adapter module that simultaneously leverages knowledge from the original CLIP and new learning knowledge derived from downstream datasets. Benefiting from the pseudo visual prompts, our method surpasses the state-of-the-art not only on clean annotated text data but also on pseudo text data generated by large language models.
\end{abstract}

\section{Introduction}
Vision and language pre-training models~\cite{CLIP:conf/icml/RadfordKHRGASAM21,BridgeTower:conf/aaai/0005WRLCD23,APT:conf/cvpr/BowmanAZTPPS23,NAIC:conf/aaai/FuSZY24, wan2024covlr,liu2023qtiah}, by learning versal knowledge representations from large-scale image-text pairs, have demonstrated remarkable generalization capabilities across many downstream tasks. This is attributed to the contrastive pre-training approach that aligns images and texts in a shared latent space. However, when data or labels are hard to collect, continuing to fine-tune the entire model often becomes challenging. Therefore, prompt tuning~\cite{CoCoOp:conf/cvpr/ZhouYL022,PPT:conf/acl/GuHLH22,Hierarchical-Prompt:conf/ijcai/WangYWD23,APVT:conf/ijcai/SunLPMZX023} has emerged as an efficient and intelligent new paradigm for adapting large models to small-scale tasks.
\begin{figure}[h]
    \centering
    \includegraphics[width=\linewidth]{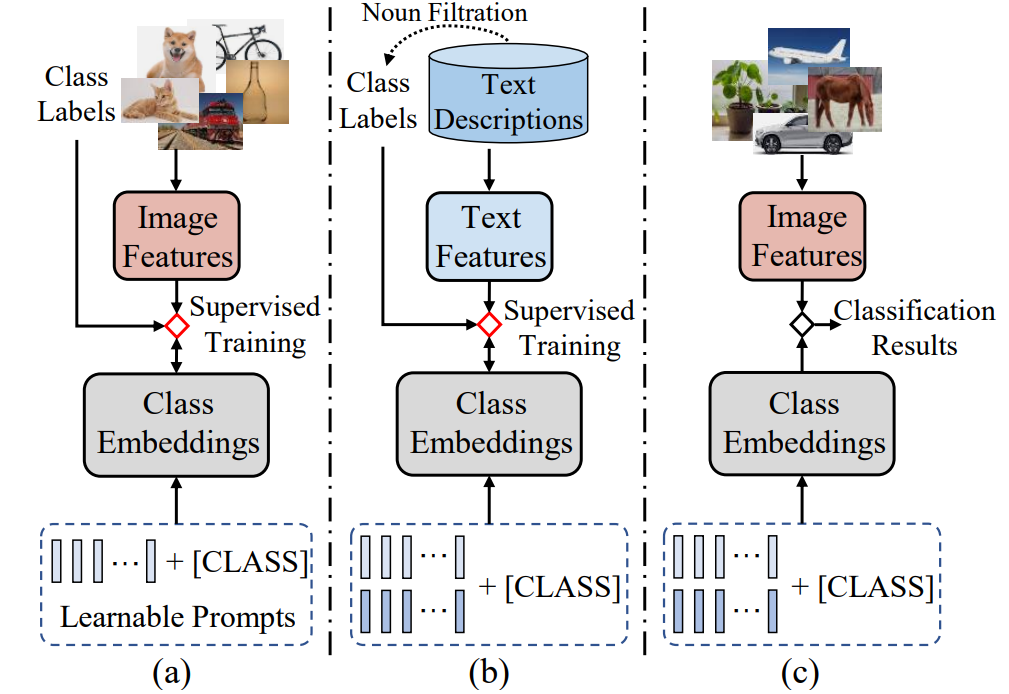}
    \caption{(a): NoCaps dataset, which always includes common objects such as animals, plants and furniture, etc. (b)(c): NICE Challenge dataset, which includes many novel visual concepts and various image types, such as famous historic, cultural and graphics, etc.}
    \label{fig:example}
\end{figure}
Existing prompt tuning methods for image recognition mainly focus on entire or partially labeled image settings. For example, both CoOp~\cite{CoOp:journals/ijcv/ZhouYLL22} and CoCoOp~\cite{CoCoOp:conf/cvpr/ZhouYL022} require annotated images for training learnable context, while DualCoOp~\cite{DualCoOp:conf/nips/SunHS22} is trained from partially labeled images for both positive and negative prompts. When visual or labeled data is limited, TAI-DPT~\cite{TAIDPT:conf/cvpr/GuoDJBGZ23} further proposes treating texts as images for prompt tuning, without seeing any image and training double-grained text prompts, and successfully applied to zero-shot multi-label image recognition. Nevertheless, learning a wide range of discriminative visual-level features for each target category is crucial for zero-shot image recognition, and relying solely on text-level labels and corresponding text encoder to store aligned visual information in text prompts is insufficient for representing the vast diversity of visual objects. For instance, categories such as car, plane, or bag encompass a wide range of shapes and attributes. This heterogeneity can easily result in text prompts overfitting to local feature spaces, hindering the ability to generalize and recognize different instances within the same category.

Given an image and a corresponding sentence, the image is processed through an image encoder into a d-dimensional vector, while the text undergoes a similar process through a text encoder. These vectors are then aligned via contrastive learning in a shared latent space. This training process can be formalized as: 
\[
<I, T, EncI, EncT> \rightarrow <\text{Aligned Shared Latent Space}>
\]
where a vision-language alignment latent space is trained on a large scale of image-text pairs. 

Considering a zero-shot scenario without any image data, this approach leverages the learned aligned latent space of CLIP~\cite{CLIP:conf/icml/RadfordKHRGASAM21}, and the text modality to reverse recover the image modality:
\[
<\text{Aligned Shared Latent Space}, T, EncI, EncT> \rightarrow <I>
\]
We refer to the recovered image \(I\) as a pseudo-visual prompt. 

For the downstream zero-shot image recognition task, as referenced by~\cite{TAIDPT:conf/cvpr/GuoDJBGZ23}, instead of designing text prompts concatenated with textual labels before the frozen text encoder, we randomly initialize an identical visual prompt of size \(H \times W \times C\) for each category without combining it with any visual or textual labels before the frozen image encoder. 

For each object, similar to~\cite{TAIDPT:conf/cvpr/GuoDJBGZ23}, we treat texts as images. Their textual labels are derived from the input texts, constructing positive samples with the visual prompts corresponding to the categories and negative samples with other categories’ prompts. Each pseudo-visual prompt is optimized via contrastive learning in CLIP’s aligned latent space on large-scale sentence data generated by large language models.

After training, the learned pseudo visual prompts, which have a similar size and channels to real images, represent all shapes and attributes of the object category with a class-specific visual prompt. We can directly compute the similarity between the class-specific visual prompt and the input image, thus performing zero-shot multi-label image recognition, which achieves performance comparable to~\cite{TAIDPT:conf/cvpr/GuoDJBGZ23}.

However, considering the weaker visual representation capability of text prompts, we further transfer the extensive visual information contained in the pseudo-visual prompt to the text prompts to enhance their visual representation capability and image recognition performance. Specifically, we perform contrastive learning between the trained pseudo-visual prompt and the text prompts to enhance the visual diversity representation capability of the text labels. Additionally, we introduce a dual-adapter module that simultaneously leverages knowledge in the original CLIP and new learning knowledge derived from downstream datasets.
\section{Related Work}
Prompt tuning~\cite{Noisy:conf/acl/MinSMHZL22,PLPrompt:conf/ijcai/YanHXLW23,DVLF:journals/chinaf/YangBGZYY23,APT:conf/cvpr/BowmanAZTPPS23,Black-box:conf/ijcai/YuCL023} has emerged as a promising technique in computer vision and natural language processing, offering a parameter-efficient way to leverage large pre-trained models. KnowPrompt~\cite{KnowPrompt:conf/www/ChenZXDYTHSC22} involved injecting knowledge into a prompt template and encoding rich semantic knowledge among entities and relations. Pro-Tuning~\cite{ProTuning:journals/corr/abs-2207-14381} learned task-specific visual prompt for downstream input images while keeping the pre-trained model frozen. 

Amidst the progress in multi-modal pre-training, researchers have explored the application of prompt tuning in the multi-modal domain\cite{yang2019comprehensive,yang2022domfn}. CoOp~\cite{CoOp:journals/ijcv/ZhouYLL22} modified the pre-trained vision-language models for image recognition tasks by employing learnable prompt context vectors. DualCoOp++~\cite{DualCoOp++:journals/pami/HuSXSS23} efficiently adapted a powerful vision-language model with partial-labeled images by introducing evidence-guided region feature aggregation and winner-take-all modules to improve spatial aggregation and inter-class interaction. These methods necessitated visual modality and textual class labels as default prerequisites in both training and testing. Consequently, TAI-DPT~\cite{TAIDPT:conf/cvpr/GuoDJBGZ23} extended this paradigm by treating texts as images for zero-shot image recognition, storing the aligned vision-language information from CLIP into text prompt without seeing any image during training.

Prompt tuning based methods can boost the performance of multi-label image classification\cite{yang2018complex,yang2019semi,yang2021cost}. However, these methods either require a large amount of labeled visual data or fail to learn the diversity of visual knowledge. In this paper, we propose a novel transferable prompt co-learning method to solve this problem by designing a pseudo-visual prompt module.
\section{Method}
\begin{figure*}[h]
    \centering
    \includegraphics[width=\textwidth]{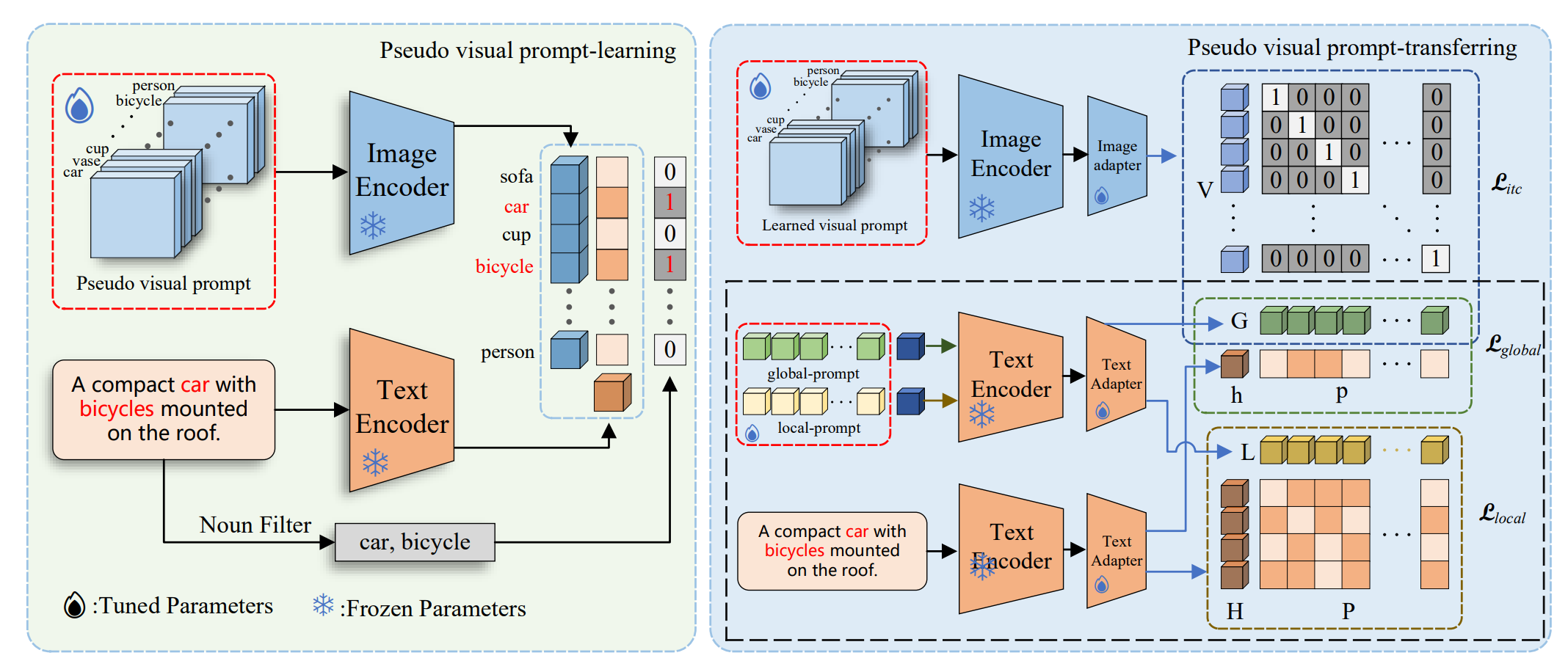}
    \caption{ Learning and transferring process of Pseudo Visual Prompt, where we use human-annotated texts or pseudo texts generated by LLM to train the prompts. (a) During learning, we design identical class-specific visual prompts for each target category. The global text feature and object visual features are obtained from the frozen CLIP image and text encoder. The corresponding cosine similarity between the embeddings is guided by the derived pseudo labels with ranking loss. (b) During transferring, we perform contrastive learning between the trained pseudo-visual prompt and the text prompts to enhance the visual diversity representation capability of the text labels. The final classification results are obtained by merging the scores of the two branches.}
    \label{fig:example}
\end{figure*}

We focus on the task of zero-shot multi-label image recognition. To begin with, we revisit the training process of TAI-DPT~\cite{TAIDPT:conf/cvpr/GuoDJBGZ23} as shown in the black box of Fig \, and highlight the disadvantages of the designed text prompts. Then, we introduce pseudo-visual prompt learning, which includes the construction of two forms of sentence data and the training of visual prompts. After learning, we adopt contrastive learning to transfer our pseudo-visual prompts to text prompts. Finally, we discuss the entire model's training objectives and inference process.
\subsection{Background}

Figure \ref{fig:example} illustrates the approach used by TAI-DPT~\cite{TAIDPT:conf/cvpr/GuoDJBGZ23}, which treats text as a proxy for images to achieve zero-shot multi-label image recognition. In this method, global (G) and local (L) class-specific embeddings are generated by concatenating double learnable text prompts with each class embedding. These concatenated embeddings are then processed through the frozen text encoder of the original CLIP model. The global text feature of the input sentence and the sequential feature of word tokens, denoted as h and H respectively, are derived using the same frozen text encoder. It's important to note that the original TAI-DPT~\cite{TAIDPT:conf/cvpr/GuoDJBGZ23} does not incorporate a text adapter module. The training objectives are formulated as follows:

\[ 
L_{\text{global}} = \sum_{i \in \{c^+\}} \sum_{j \in \{c^-\}} \max(0, m - p_i + p_j)
\]
\[ 
L_{\text{local}} = \sum_{i \in \{c^+\}} \sum_{j \in \{c^-\}} \max(0, m - p'_i + p'_j)
\]

In these equations, \( c^+ \) and \( c^- \) refer to the filtered positive object classes and the remaining negative classes identified by a noun filter (e.g., car, bicycle). The variable \( p \) signifies the global similarity between the global class embedding \( G \) and the global sentence feature \( h \), whereas \( p' \) denotes the aggregated local similarities between the local class embedding \( L \) and the sequential word feature \( H \). The margin \( m \) ensures that the similarity with positive classes is significantly higher than with negative classes. During the testing phase, images take the place of texts, where \( h \) and \( H \) represent the global feature and local patch feature of the image, respectively. The final classification score is obtained by combining \( p \) and \( p' \).

\subsection{Pseudo Visual Prompts Learning}

Accurately learning extensive and diverse visual representations for each object category is crucial for image recognition. Traditional text prompts and text encoders capture visual information aligned solely with specific textual categories in the CLIP alignment space, limiting their ability to fully represent the feature space of a category (e.g., capturing only the visual elements associated with the label "car"). To address this limitation, we introduce pseudo-visual prompts, which construct class-specific visual prompts for each category. These prompts utilize the image encoder, text encoder, and shared latent space of CLIP to optimize and learn the generic visual features of each category. Additionally, we transfer the visual representations from the pseudo visual prompts to text prompts to enhance their visual representation capabilities and improve image recognition performance.

\subsubsection{Construction of Text Training Data and Textual Labels}

Given that CLIP aligns texts and images in a shared latent space, training image recognition tasks with text data instead of images is feasible. This necessitates that the sentence descriptions used as training data meet the following criteria:
\begin{enumerate}
    \item Each sentence should include an appropriate number of textual labels, with a balanced distribution to avoid long-tail issues.
    \item The sentence content should richly describe a natural image scene, akin to real image descriptions.
\end{enumerate}

To obtain such text descriptions, we employ two methods: human-annotated clean data and pseudo data generated by large language models. For clean data, we utilize public image caption datasets such as MS-COCO and object detection datasets like localized narratives. For pseudo-text data, we create a text prompt template and use a large language model to generate descriptions automatically.

For a target category set \( C = \{c_1, c_2, c_3, \ldots, c_n\} \), where \( n \) represents the number of categories and \( c_i \) denotes a specific class, we design a query template as follows: "Create a sentence to describe a photo. Requirements: Each sentence should be less than 15 words and include keywords: \{$c_1$, $c_2$, \ldots, $c_i$\}," where \(\{c_1, c_2, \ldots, c_i\} \subseteq C\). This query is then input into ChatGLM to automatically generate pseudo text descriptions. Due to the inherent variability of large language models, we re-input the generated data into ChatGLM to verify the reasonableness of the sentences, retaining only the most reliable training data. For word-level filtered labels in the input sentences, we use NLTK to perform noun filtering, following the TAI-DPT setup. More details on the query template and noun filtering process are provided in the Appendix.

\subsubsection{Pseudo Visual Prompts}

Unlike CoOp, dualCoOp, and TAI-DPT, which design text prompts for textual labels, we introduce class-specific visual prompts without combining them with explicit visual or textual labels on the image side. For example, text prompts must explicitly concatenate with textual labels like "cat" or "dog." However, the image modality encompasses a wide range of diverse shapes and attributes. Therefore, we design a class-specific visual prompt for each category to learn and store the unique visual information of that category. Specifically, for a batch containing \( B \) input sentences, the pseudo visual prompt is defined as:

\[
\text{PVP} = [P_1, P_2, P_3, \ldots, P_n]
\]

where \( P_i \in \mathbb{R}^{H \times W \times 3} \) represents the class-specific pseudo visual prompt for the \( i \)-th class, and \( N \) is the number of categories. The size of PVP is independent of batch size and depends only on the number of target categories \( N \).

We optimize the pseudo visual prompts through the aligned shared space of CLIP, using both the text encoder and the image encoder, to learn the diverse and comprehensive visual information for each class. The training process is formalized as follows:

\[
\langle \text{Aligned Space}, T, \text{EncT}, \text{EncI} \rangle \rightarrow \langle \text{PVP} \rangle
\]

where \( T \) represents the collected sentence data from public datasets or generated automatically. For the input sentence \( T \), we follow CLIP's method to obtain the global text feature by projecting the feature of the last \texttt{<EOS>} token, and the global visual feature for each category of PVP is obtained by visual attention pooling. Thus, we have:

\[
F_T = \text{EncT}(T)
\]
\[
F_I = \text{EncI}(\text{PVP})
\]

Here, EncT and EncI are the text encoder and image encoder of the original CLIP, respectively. \( F_T \in \mathbb{R}^{B \times D} \) represents the extracted global text features of a batch, while \( F_I \in \mathbb{R}^{N \times D} \) represents the global visual features of \( N \) pseudo visual prompts. For a specific sentence \( t \), the similarity between the sentence \( t \) and the pseudo visual prompts is computed by:

\[
s_n = \langle F_t, F_n \rangle, \quad n = 1, 2, 3, \ldots, N
\]

where \( F_t \) and \( F_n \in \mathbb{R}^{D} \) are the global text feature of sentence \( t \) and the global visual feature of the \( n \)-th category. We then perform noun filtering to obtain the positive labels contained in the sentence (e.g., car, bicycle), while other categories are denoted as negative labels. The global text feature and pseudo visual prompts corresponding to the positive labels form positive pairs, whereas the global text feature and the negative labels form negative pairs. Following TAI-DPT, we use the ranking loss to measure the discrepancy between similarity scores and text labels instead of binary cross-entropy loss:

\[
L_{\text{PVP}} = \sum_{i \in \{c^+\}} \sum_{j \in \{c^-\}} \max(0, m - s_i + s_j)
\]

where \( c^+ \) and \( c^- \) are positive and negative labels, \( s_i \) and \( s_j \) are the positive and negative pair similarities described in Eq. (5), and \( m \) is the margin that ensures the positive pair score is at least \( m \) higher than the negative pair score. During training, we freeze the text encoder and image encoder and optimize only the pseudo visual prompt.
\subsection{Pseudo Visual Prompts Transferring}

Upon completing the pseudo visual prompt learning, the class-specific visual information for each category is stored in the corresponding class-specific prompt, enabling zero-shot multi-label image recognition by replacing \( F_t \) in Eq. (5) with the global image feature. Our results indicate that the pseudo-visual prompts achieve competitive image recognition performance compared to TAI-DPT. Moreover, the text prompts in TAI-DPT are insufficient to capture the visual diversity of object categories. Therefore, we propose transferring the visual information encapsulated in the pseudo-visual prompts to the global text prompts via contrastive learning to enhance their visual representation capability.

Specifically, the pseudo-visual prompt and the text global prompt are represented as follows:

\[
\text{PVP} = [P_1, P_2, P_3, \ldots, P_n]
\]
\[
\text{TGP} = [T_1, T_2, T_3, \ldots, T_M, G_i]
\]

where \( G_i \) denotes the word embedding of the \( i \)-th class, and for \( j \in \{1, 2, \ldots, M\} \), \( T_j \) is a learnable word embedding. We also introduce a dual adapter consisting of two MLP layers to leverage both the original CLIP's knowledge and the new knowledge from downstream datasets. Thus, we have:

\[
F_T = \text{MLP}(\text{ReLU}(\text{MLP}(\text{EncT}(\text{TGP}))))
\]
\[
F_I = \text{MLP}(\text{ReLU}(\text{MLP}(\text{EncT}(\text{PVP}))))
\]

where \( F_T, F_I \in \mathbb{R}^{N \times D} \) are the global visual prompt features and global text prompt features of \( N \) categories, respectively. The similarity matrix can then be obtained by \( F_T F_I^T \), where the diagonal elements are 1 and the off-diagonal elements are 0. Note that the size of the similarity matrix is batch-size agnostic and is always \( N \times N \). The contrastive loss is computed as follows:

\[
L_{\text{tpc}} = \sum_{i=1}^{N} \text{CE}(w_{i,\text{soft}}) + \text{CE}(w_{i,\text{tpc}})
\]

where \( w_{i,\text{soft}} \) and \( w_{i,\text{tpc}} \) denote the softmax-normalized text-to-visual prompt and visual-to-text prompt similarities with category size \( N \) and temperature scale parameter \( \tau \). \( \text{CE} \) denotes the cross-entropy loss. The total training loss is then given by \( L = L_{\text{tpc}} + L_{\text{global}} + L_{\text{local}} \), where \( L_{\text{global}} \) and \( L_{\text{local}} \) are the loss terms for global text embedding and local text tokens described in Section 3.1, respectively.

\subsection{Model Inference}

The pseudo-visual prompt aims to learn diverse and comprehensive visual information and transfer it to the text prompt to enhance visual representation capability. During testing, we can directly remove the pseudo visual prompt and maintain the same testing settings as TAI-DPT without adding any extra inference cost. Specifically, we replace the input from text descriptions with images, then compute global and local classification scores using class embeddings generated by the global and local prompts via cosine similarity. The final classification result is obtained by fusing the global and local classification scores.
\section{Experiments}
\subsection{Metric}
We use Mean Average Precision (mAP) to evaluate multi-label classification accuracy. The mAP is calculated as follows:
\[
mAP = \frac{1}{C} \sum_{i=1}^{C} AP_i
\]
where \( C \) is the total number of classes and \( AP_i \) is the Average Precision for the \( i \)-th class. The Average Precision (AP) is defined as:
\[
AP = \frac{1}{m} \sum_{k=1}^{m} \frac{Positives@k}{Total@k}
\]
where \( m \) is the number of relevant items, \( Positives@k \) represents the number of true positives up to the \( k \)-th position, and \( Total@k \) is the total number of items up to the \( k \)-th position.
\subsection{Results}
Table \ref{tab:results} presents the mean Average Precision (mAP) results for various methods applied to multi-label classification. Below is an in-depth analysis of the performance of each method.

\begin{table}[h]
\centering
\begin{tabular}{c|c}
\hline
\textbf{Method}                   & \textbf{mAP}  \\ \hline
Baseline                        & 64.68         \\ \hline
Data Augmentation               & 65.35         \\ \hline
Reasonableness Check            & 66.14         \\ \hline
Text Noise                      & 67.82         \\ \hline
Pseudo Visual Prompt Pre-training       & 71.31 (\textcolor{red}{+3.49}) \\ \hline
Knowledge Transfer              & 72.39         \\ \hline
Dual-Adapter                    & 73.13         \\ \hline
TTA                             & 75.07         \\ \hline
\end{tabular}
\caption{mAP Results of Different Methods for Multi-label Classification}
\label{tab:results}
\end{table}

The baseline method achieves an mAP of 64.68, serving as the reference point for evaluating subsequent methods. Data augmentation slightly improves the mAP to 65.35, demonstrating the benefit of leveraging additional generated data to enhance model generalization. Reasonableness checks further enhance the mAP to 66.14 by ensuring the plausibility and relevance of the training data. Introducing text noise significantly increases the mAP to 67.82, indicating that robustness to noisy and imperfect data is beneficial.

The pseudo visual prompt pre-training method markedly boosts the mAP to 71.31, with an improvement of +3.49 over the baseline. This substantial increase highlights the efficacy of incorporating pseudo-visual prompts, which enhance the model’s ability to capture diverse and extensive visual information. Knowledge transfer further improves the mAP to 72.39, suggesting the advantage of leveraging additional knowledge from related tasks or datasets to improve generalization to the target task.

The dual-adapter method achieves an mAP of 73.13, effectively combining knowledge from the original CLIP model and new downstream datasets. This approach results in enhanced performance by integrating multiple sources of knowledge. Test-time augmentation (TTA) provides the highest mAP of 75.07, likely due to the averaging of predictions over multiple augmented versions of the test data, which reduces variance and enhances robustness.

In summary, the results clearly indicate that the pseudo visual prompt pre-training method significantly improves model performance in multi-label classification tasks. Further enhancements are observed with knowledge transfer, dual-adapter, and test-time augmentation methods. Each method contributes to better capturing and utilizing visual and textual information, leading to higher mAP scores.
{\small
\bibliographystyle{ieee}
\bibliography{egpaper_final}
}

\end{document}